\documentclass[letterpaper, 10 pt, journal, twoside]{IEEEtran}  

\IEEEoverridecommandlockouts                              

\usepackage{times}
\usepackage[pdftex]{graphicx}
\usepackage{subfigure} 
\usepackage{booktabs}
\usepackage{epsfig} 
\usepackage{amsmath} 
\usepackage{amssymb}  
\usepackage{breqn}
\usepackage{cite}
\usepackage{bm}
\usepackage{multicol, blindtext}
\usepackage{float}
\usepackage{siunitx}
\usepackage{xcolor}
\usepackage{soul}
\usepackage{url}
\usepackage{lscape} 
\usepackage{makecell}
\usepackage{pdfpages}

\DeclareGraphicsExtensions{.png,.pdf}
\usepackage{multirow}
\usepackage{hyperref}
\usepackage{color,linegoal}
\renewcommand\hl[1]{#1} 
\begin{document}
\title{
TraKDis: A Transformer-based Knowledge Distillation Approach for Visual Reinforcement Learning with Application to Cloth Manipulation

}

\author{~Wei~Chen$^{1}$ and Nicolas~Rojas$^{1}$~\IEEEmembership{Member,~IEEE}%
\thanks{Manuscript received: September 1, 2023; Revised: November 22, 2023; Accepted: January 18, 2024.}
\thanks{This paper was recommended for publication by Editor Jens Kober upon evaluation of the Associate Editor and Reviewers' comments.
This work was supported in part by the China Scholarship Council and the Dyson School of Design Engineering, Imperial College London.} 
\thanks{$^{1}$Wei Chen and $^{1}$Nicolas~Rojas are with the REDS Lab, Dyson School of Design Engineering, Imperial College London, 25 Exhibition Road, London, SW7 2DB, UK
        {\tt\footnotesize (w.chen21, n.rojas)@imperial.ac.uk}}%

\thanks{Digital Object Identifier (DOI): see top of this page.}}

\markboth{IEEE Robotics and Automation Letters. Preprint Version. Accepted January, 2024}
{Chen \MakeLowercase{\textit{et al.}}: TraKDis: A Transformer-based Knowledge Distillation Approach for Visual Reinforcement Learning}

\maketitle

\begin{abstract}

Approaching robotic cloth manipulation using reinforcement learning based on visual feedback is appealing as robot perception and control can be learned simultaneously. However, major challenges result due to the intricate dynamics of cloth and the high dimensionality of the corresponding states, what shadows the practicality of the idea. To tackle these issues, we propose \textit{TraKDis}, a novel \textit{Tra}nsformer-based \textit{K}nowledge \textit{Dis}tillation approach that decomposes the visual reinforcement learning problem into two distinct stages. In the first stage, a privileged agent is trained, which possesses complete knowledge of the cloth state information. This privileged agent acts as a teacher, providing valuable guidance and training signals for subsequent stages. The second stage involves a knowledge distillation procedure, where the knowledge acquired by the privileged agent is transferred to a vision-based agent by leveraging pre-trained state estimation and weight initialization. \textit{TraKDis} demonstrates better performance when compared to state-of-the-art RL techniques, showing a higher performance of 21.9\%,  {13.8\%}, and  {8.3\%} in cloth folding tasks in simulation. Furthermore, to validate robustness, we evaluate the agent in a noisy environment; the results indicate its ability to handle and adapt to environmental uncertainties effectively. Real robot experiments are also conducted to showcase the efficiency of our method in real-world scenarios. 




\end{abstract}

\section{Introduction}





\IEEEPARstart{M}{anipulating} deformable objects, especially cloth and other fabrics, with robotic technology is a considerable challenge that has far-reaching and widespread applications in domestic, medical, and industrial settings. Important tasks such as assistive dressing, fabric manufacturing automation, and household cleaning tasks all require robust, delicate, and precise manipulation of cloth to achieve their goals~\cite{doi:10.1126/scirobotics.abo7229}. While the manipulation of rigid-body objects is relatively well-developed, that of cloth remains in its infancy.


Reinforcement learning (RL) is a popular control tool for robotic manipulation tasks that has been utilised in cloth manipulation successfully~\cite{matas2018sim, lin2021softgym,9196659,wu2019learning}. However, RL agents are typically trained using cloth state information, such as the exact particle positions of keypoints on the cloth. This allows the agent to achieve satisfactory performance in simulated environments; however, obtaining such state information in the real world is particularly challenging. Rather than relying on exact object states, some works infer them from RGB image data in a visual manipulation policy~\cite{lin2021softgym, salhotra2022learning}. However, while this is often successful in rigid-object manipulation~\cite{levine2016end, levine2018learning}, cloth and fabric exhibit high self-occlusion and are often lacking in tracking observable features such as edges and corners, making visual manipulation difficult. Further still, visual data such as RGB image data always presents higher dimensionality than state information data, which makes directly training a visual RL policy more challenging.


Most of the existing visual RL-based works for cloth manipulation employ an asymmetric actor-critic architecture for their policy training. This is implemented by using high-dimension observation (RGB images) and low-dimension environment states as the input for the actor and critic, respectively~\cite{pinto2017asymmetric}. Although this asymmetric architecture has been applied in rigid-body and deformable object manipulation~\cite{liu2022distilling,lin2021softgym, salhotra2022learning}, the utilization of actor-critic also comes with the limitation that all decisions are made based on the Markov assumption, which may cause robots to lose historical information~\cite{10160946,ramakrishnan2021exploration}. 

\begin{figure*}[t!]
    \centering
    \vspace{4mm}
    \includegraphics[width=0.75\textwidth]{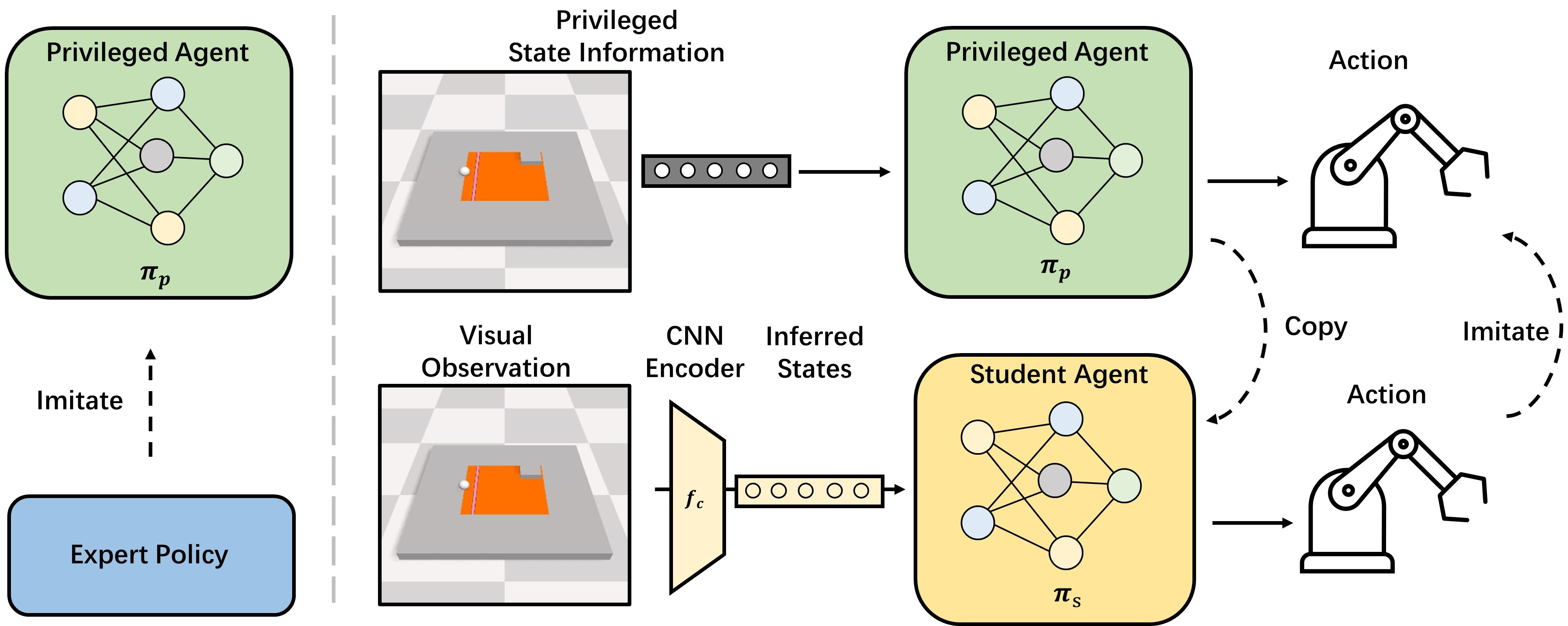}
    \caption{Pipeline of \textit{TraKDis} for training visual RL agents for cloth manipulation. A privilege agent is trained to imitate the expert policy with privileged state information (cloth particle locations) to obtain a robust performance. Then, a student agent that can only access partial observation (RGB images) is trained to imitate the privileged agent. By leveraging a CNN encoder and weight copy, the student policy can realize an enhanced performance with limited observation information.}
    \label{fig:pipeline figure}
    \vspace{-10pt}
\end{figure*}

In a highly dynamic environment such as cloth manipulation, the current state of cloth may depend not only on the last state but also on a sequence of past states. The loss of historical information may influence the controllability and observability of the overall control agent. To address this issue, in contrast to previous works\cite{salhotra2022learning, lin2021softgym}, we adopt a transformer-based RL approach for the learning framework. We argue that the adoption of the transformer model yields twofold benefits. First, the transformer architecture can process information with a large capacity and a long memory, which holds more significant promise for accurately predicting and understanding complex trajectory sequences~\cite{ding2023learning}. Second, transformers excel at capturing contextual relationships, which enhance knowledge transfer by understanding data nuances and dependencies. Pre-training on a large dataset enables recognition and comprehension of contextual patterns, aiding fine-tuning specific tasks~\cite{humeau2019poly}. 



This work then proposes \textit{TraKDis}, a novel \textbf{Tra}nsformer-based \textbf{K}nowledge \textbf{Dis}tillation (KD) approach for learning visual cloth manipulation. Our goal here is to solve the highly dynamic cloth manipulation tasks with only RGB image input. Unlike the asymmetric input of RGB images and states in actor-critic approaches~\cite{salhotra2022learning}, we apply KD for the learning of vision-based tasks. While most KD procedures focus on compressing multiple teachers' knowledge into one student model, \textit{TraKDis} can be considered as a special \textit{one-to-one} case~\cite{zhu2023transfer}.
As shown in Fig.\ref{fig:pipeline figure}, \textit{TraKDis} decomposes the learning of the vision agent into two stages. First, we train a privileged agent that takes the privileged cloth state information as input. A novel KD procedure is then conducted to distil the knowledge from the privileged agent to the vision-based (student) agent. Our proposed KD approach consists of two major components: a state estimation encoder and the weight transfer. Specifically, the pre-trained CNN state estimation encoder is applied as the \textit{prior} to project the high-dimensional images into estimated low-dimensional states. This aims to reduce the domain gap between visual observation and states during the distillation. 
Since the two models share the same architecture, we also initialize the student policy by weight transfer from the privilege agent at the beginning of distillation, which shows significant improvement in facilitating knowledge distillation.

 We highlight our key contribution as follows: (1) \textit{TraKDis}: a novel transformer-based knowledge distillation framework for learning visual cloth manipulation tasks. (2) We propose a novel approach for the knowledge distillation procedure by leveraging a state estimation encoder and pre-trained weight. The student model (image) can leverage the teacher model (states) knowledge, significantly increasing the model's performance and training efficiency.
 (3) With the proposed learning framework, \textit{TraKDis} outperforms other state-of-the-art (SOTA) baselines in several comparison experiments.

\section{Related Works}
\subsection{Cloth Manipulation by Data-driven Methods}
Due to the characteristics of high elasticity, low controllability, and complex dynamics, cloth manipulation remains a challenge in the field of robotic manipulation\cite{{9097275,8957044}}. In recent years, some data-driven methods approaches have gained popularity. Previous works tried to use deep learning (DL)-based methods to detect key cloth regions for downstream tasks such as bed-making, cloth grasping, cloth flattening and garment hanging~\cite{seita_bedmake_2019, qian2020cloth, ha2021flingbot, xu2022dextairity, chen2023learning, Wang2023One}.

 By learning from trial and error, RL and RL combined with Imitation Learning (IL) methods have made significant progress in deformable object manipulation, especially cloth manipulation tasks. Using the combination of deep deterministic policy gradient (DDPG), Hindsight Experience Replay (HER) and expert demonstration, \cite{9196659} tries to learn the dynamic cloth folding tasks. However, one preliminary of this approach is the access to the state information of cloth during the whole manipulation procedure. In the real world, however, extracting such ground truth state information is not always available for a robot because of the large configuration space of deformable objects. On the other hand, rather than state information, it is more feasible for a robot to use high-dimensional observations such as RGB images as inputs to its control policy~\cite{lin2021softgym}.

 Due to the deformable nature of cloth and the typically high-dimensional configuration parameters, it is extremely challenging to infer the cloth states from only visual input. This makes vision-based cloth manipulation more difficult. With some recent progress in  RL, CURL~\cite{laskin2020curl} and PlaNet~\cite{hafner2019learning} are proposed for the training of image-based agents.
 Although several image-based RL methods, including CURL and PlaNet have been used as the benchmark to test the performance of visual cloth manipulation tasks in~\cite{lin2021softgym}, the results are far below the optimal performance (state-based agent) on many tasks. Furthermore, by combining the expert demonstration and RL,~\cite{salhotra2022learning} has adopted an asymmetric architecture for Learning from Demonstration (LfD) methods, where the actor acts based solely on environment images while the critics learn from both image and state information. However, this still results in a significant gap when the agent performs tasks with only image observation compared to state input.

\subsection{Transformer}
Benefiting from the attention mechanism and its significant advantages of sequence modelling, transformers are initially applied in the field of Natural Language Processing (NLP)~\cite{vaswani2017attention}. Being attributed to its high-quality global contextual learning and friendly parallel computation, transformer architecture has emerged as a high-capacity learning framework over traditional sequence prediction methods such as Recurrent Neural Networks (RNNs) or Long Short-Term Memory Networks (LSTMs).

With some breakthroughs in transformers, the application area of transformers also extends to vision and reinforcement learning (RL)~\cite{dosovitskiy2020image,chen2021decision,zheng2022online}. For instance, \cite{dosovitskiy2020image} proposed the vision transformer (ViT) which shows better accuracy over the traditional CNN with a large training dataset. 
Furthermore, researchers have also utilized transformer architecture in decision-making agents, proposing the decision transformer (DT) model, which re-frames the sequence modelling problem as an action prediction problem based on a series of historical states and reward-to-go\cite{chen2021decision}. Training in a supervised fashion, DT tries to minimize the error between predicted and ground-truth actions by inputting past state and reward-to-go sequences. Some recent work has shown DT has superiority over many state-of-the-art offline RL algorithms~\cite{chen2021decision}. 
Furthermore, more variants of DT have been proposed to extend the DT model to the online training scenario~\cite{chen2021decision} and multi-agent scenario~\cite{lee2022multi}.

\subsection{Knowledge Distillation}
Knowledge distillation is a transfer learning (TL) approach that aims to transfer knowledge from the teacher model to the student model~\cite{zhu2023transfer}.   
The original idea of knowledge distillation (KD) is to transfer knowledge from a large and complex model (the teacher) to a small and simplified model (the student)~\cite{hinton2015distilling}. This transfer is usually achieved by minimizing the Kullback–Leibler (KL) divergence of teacher and student outputs. 

Some recent work has integrated state estimation and TL in robotics for locomotion control of legged robots. 
\hl{By integrating a Temporal-Convolutional Neural Network (T-CNN) or temporal-MLP, temporal information can be applied to enhance state estimation for the learning of the student policy in these methods \mbox{\cite{kumar2021rma, lee2020learning}}. In contrast, \textit{TraKDis} utilizes the DT architecture for temporal decision-making, using the state input estimated by the CNN encoder.}
 

\begin{figure}[t!]
    \centering
    \vspace{4mm}
    \includegraphics[width=0.98\columnwidth]{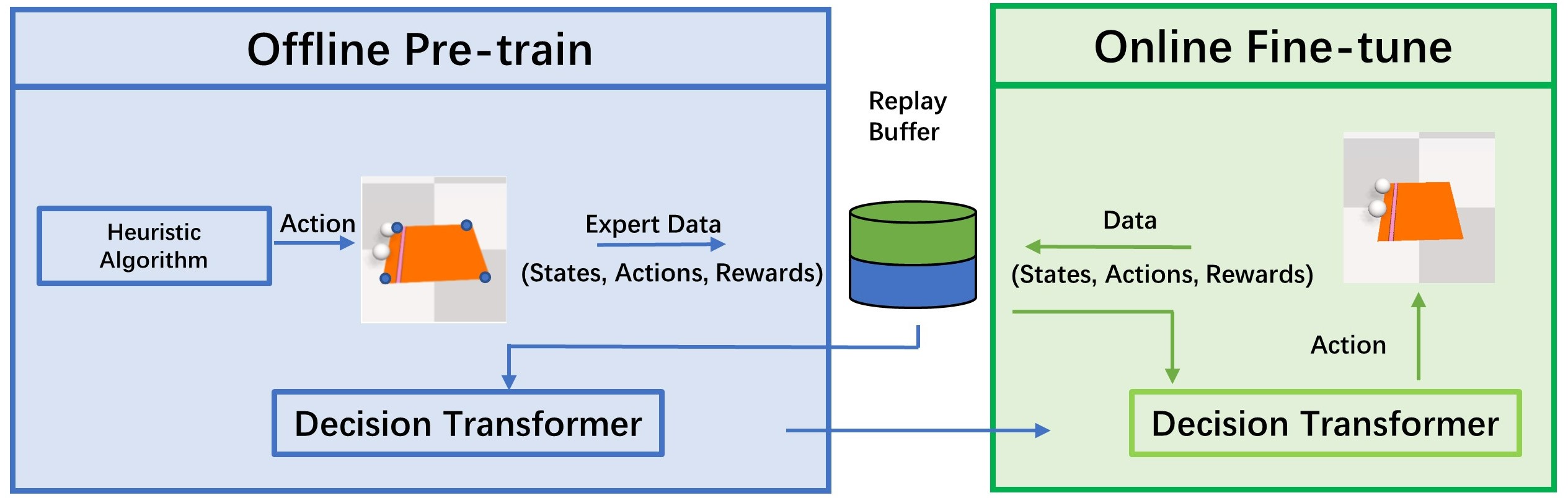}
    \caption{Model Training of the privileged agent. We first conduct offline training using the expert data collected from the human-designed heuristic algorithm. An online fine-tuning is then applied for the trained model to obtain a better performance} 
    \label{fig:model_training}
    \vspace{-10pt}
\end{figure}

\section{Methods}

\subsection{Preliminary}

We formulate the cloth manipulation environment as a partially observable Markov decision process (POMDP). The tuple of ${<\mathcal{A},\mathcal{S},\mathcal{O},\mathcal{P},\mathcal{R}>}$ are used to describe the POMDP, which consists of actions $a \in \mathcal{A}$, states $s \in \mathcal{S}$, observations $o \in \mathcal{O}$, transition functions $p \in \mathcal{P}$ and rewards $r \in \mathcal{R}$.  At each timestep $t$, the RL agent receives an observation $o_t$ and generates an action $a_t$. The environment then updates the applied action based on the transition function $p(s_t+1|s_t, a_t)$ and assigns a reward to the agent. The goal of the agent is to maximize the cumulative discounted rewards $R_t=\sum_{t=k}^{H}\gamma^{t}r_t$, where $H$, $k$, and $\gamma$ are the horizon length, current timestep, and the discount factor.

\subsection{RL by Decision Transformer}
 {While many single-step, forward methods are limited by error accumulation issues, it has been shown that the transformer architecture significantly improves error accumulation \mbox{\cite{janner2021offline}}. To leverage this, the DT model\mbox{\cite{chen2021decision, zheng2022online}} is adopted in this work as depicted in Fig. \mbox{\ref{fig:teacher student}}.}  To process a sequence of trajectory $\{\hat{R_1}, s_1, a_1,...,\hat{R_T}, s_T, a_T\}$, the DT model treats the sequence data as $tokens$ and embeds them into $\{ z_i\}_{i=1}^{3T}$ with position embedding ($\hat{R}$ refers to return-to-go~\cite{chen2021decision}). Specifically, the $i$-th token, being linearly projected to query $q_i$, key $k_i$, value $v_i$, is then fed into the following self-attention layers. The $i$-th  output of the self-attention layer is computed by: 
$$
z_i=\sum_{j=1}^n \operatorname{softmax}(\left\{\left\langle q_i, k_{j^{\prime}}\right\rangle\right\}_{j^{\prime}=1}^n)_j\cdot v_j
$$
Note that the DT model only processes token $j\in[1,i]$ to predict the following action. At timestep $t$, the DT model predicts the action using tokens from the past $K$ timesteps, where $K$ is the hyperparameter referred to as the $context\\length$ here. 
Since the training of DT is supervised, we apply a cross-entropy loss $\mathcal{L}_{CE}$ to maximize the likelihood of predicted action and ground-truth action from an expert policy. For the online fine-tuning, we add an entropy loss $\mathcal{L}_{en}$ with empirical weight $\lambda$ to enhance the exploration ability of our model further ($\lambda=0.1$)~\cite{zheng2022online}. The final loss is then obtained by:
$$
 \mathcal{L} = \mathcal{L}_{CE} - \lambda\mathcal{L}_{en} 
$$

We train our privileged agent from the expert dataset with access to online fine-tuning, as shown in Fig. \ref{fig:model_training}. Similar to \cite{salhotra2022learning}, we first collect offline datasets by a heuristic expert solution that has access to the full dynamics and states of the cloth to train our privileged DT model. However, the privileged DT model only trained with offline data often has a sub-optimal performance due to the pre-collected dataset distribution. Online fine-tuning is then applied to improve our model performance further\cite{zheng2022online}. This is implemented by gradually replacing samples from the training buffer with new data collected by the trained agent, as shown in Fig.  \ref{fig:model_training}.

\subsection{Knowledge Distillation for Learning Visual Control Agent}

\subsubsection{Pre-trained CNN Encoder for State Estimation}
\begin{figure}[t]
    \centering
     \vspace{4mm}
    \includegraphics[width=0.98\columnwidth]{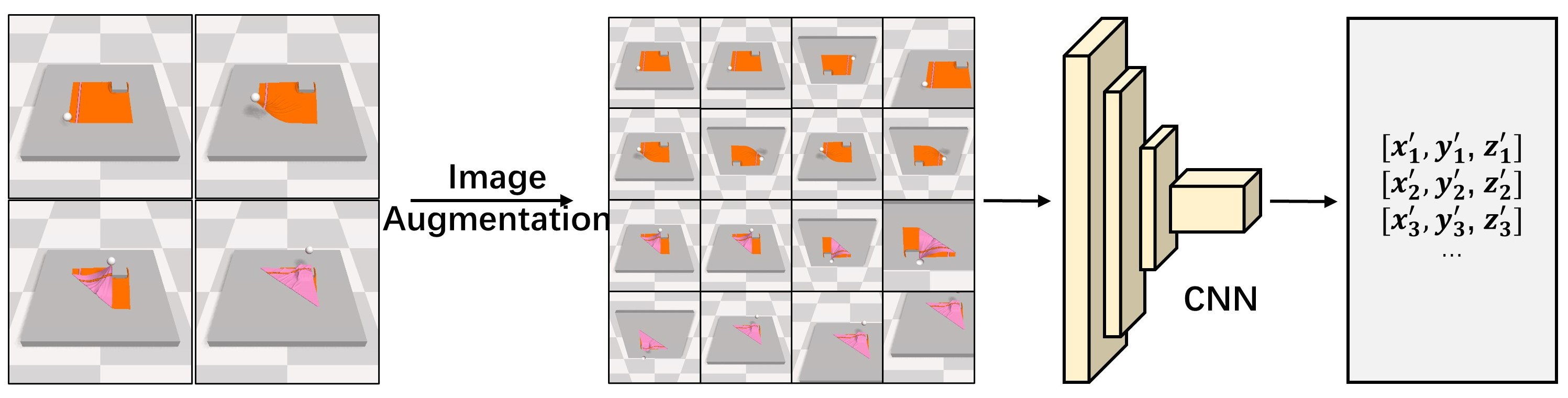}
    \caption{CNN Model: We design a CNN encoder to estimate the state information from image observation. Image augmentation is applied to improve the robustness of state estimation.}
    \label{fig:cnn_training}
    \vspace{-6mm}
\end{figure}
Due to the large domain gap between state and visual observation, learning cloth manipulation directly from visual observation is particularly challenging in RL. We design a CNN encoder $F_c$ to encode the observed images for the downstream knowledge distillation framework to minimise the gap between visual and state observation.
Specifically, our designed CNN takes an RGB image $I_t$ as the input and output estimated state representation of the target cloth $S_t^{'}$. Since simulation can give access to the state information of cloth, we can easily collect datasets without any manual labelling. In addition, recent work has also shown classic image augmentation techniques can be beneficial to visual RL training~\cite{dosovitskiy2020image,laskin2020reinforcement}. Therefore, we apply image augmentation techniques such as flipping and cropping to improve the performance of the RL agent. Being different from other methods, we only apply image augmentation in the training of the CNN encoder. Instead of augmenting the data in the RL agent training stage, we aim to implicitly improve visual RL agent performance by enhancing the robustness of the pre-trained CNN encoder.

\subsubsection{Knowledge Distillation via Weight Initialization}
\begin{figure*}[t!]
    \centering
    \vspace{4mm}
    \includegraphics[width=0.80\textwidth]{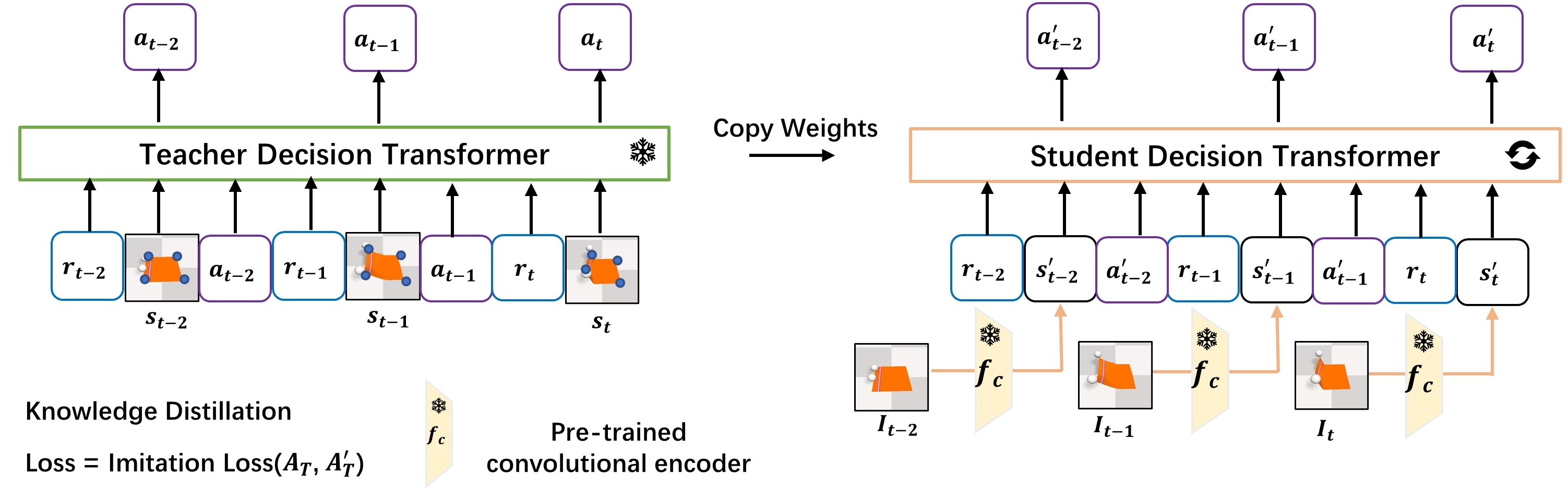}
    \caption{The figure demonstrates the overview of our knowledge distillation procedure. We run the teacher and student policy simultaneously for knowledge distillation. The teacher policy has access to all the state dynamics of the cloth and thus performs better. The student policy, which receives image inputs, is trained to imitate the actions of the teacher policy. By using a pre-trained CNN, the image can be estimated to encoded input $s_{t}$. Since the student agent and privileged agent have the same architecture, we initialize the weight of the student policy by copying the weights from the teacher's policy, which aids the knowledge distillation process. The parameters of CNN and teacher policy are frozen during training. Only the student policy is updated via the imitation loss.}
    \label{fig:teacher student}
    \vspace{-10pt}
\end{figure*}
Without any prior knowledge, training a vision-based RL is rather challenging due to the partial observability of high dynamic cloth states~\cite{salhotra2022learning}. Therefore, we adopt a KD approach for the learning of vision-based agents.
To facilitate this knowledge distillation, we propose an efficient weight initialization-based knowledge distillation framework. This framework aims to train a student policy that can imitate a privilege policy.

The student model employs a similar architecture as the privileged model. The only difference between the student model and the privileged model is that the student policy takes RGB images as input to predict actions. 
Leveraging pre-trained weight and fine-tuning has shown great potential in the transfer learning of sequence modelling of the Large Language Model (LLM). With the pre-trained weight, the fine-tuned model can obtain not only a boosted performance but also a higher training efficiency~\cite{devlin2018bert}.
Inspired by that, we initialize the student model by copying the weight of the privileged model to fully leverage the knowledge obtained by the privileged agent.
During the distillation procedure, We first apply the pre-trained CNN encoder to embed the high-dimensional image $I_t$ into low-dimension estimated states $S^'_t$. The estimated states and reward-to-go are then used to predict the action $A^'_t$.
As shown in Fig.\ref{fig:teacher student}, with the weight of the encoder and the teacher models frozen, only student policy is updated by minimizing the imitation loss of actions generated by the student and privileged models:
\vspace{-1pt}
\begin{gather*} 
{s}^{'}_{t} = {F_c}({I}_{t}), \quad
 \mathcal{L}_{distill} =\sum_{t=0}^T \mathcal{L}_{CE}({a}_{t},{a}^{'}_{t}) \\
    ({a}_{0},{a}_{1},...,{a}_{T}) \sim \mathcal{DT}_{priv}({a}_{0},{s}_{0},{R}_{0},...,{a}_{T},{s}_{T},{R}_{T}) \\
 ({a}^{'}_{0},{a}^{'}_{1},...,{a}^{'}_{T}) \sim \mathcal{DT}_{stud}({a}_{0}^{'},{s}^{'}_{0},{R}_0,...{a}_{T}^{'},{s}^{'}_{T},{R}^{'}_{T})  
\end{gather*}
where $({a}_{t},{s}_{t},{R}_{t})$ is the state, action and reward-to-go pair in the teacher's trajectories, ${F_c}$ refers to the pre-trained CNN encoder.

\section{Experiment}
In this section, we present several experiments to validate the effectiveness of our proposed method. We first provide the implementation details and task setup for our approach. We evaluate our method over these tasks to compare results with other SOTA algorithms. Besides, we conduct an ablation study to test the necessity of each component of our method. Finally, we demonstrate our model's capability to handle noisy state estimation and adapt accordingly. Real-world demonstrations are also conducted in this section.

\subsection{Tasks and Implementation Details}

We evaluate \textit{TraKDis} on three vision-based cloth manipulation tasks. All the simulation experiments are conducted in the SoftGym suite.~\cite{lin2021softgym}. For the training of privileged agent training, we use a reduced-state provided by SoftGym. For the student agent, we adopt the observation described by a 32x32x3 RGB environment image demonstrating the end-effector and target cloth. The variation of cloth deformation property is also applied for proper domain randomization of each task.

\begin{itemize}
\item
\textit{Cloth Fold Image}:
The goal of the task is to fold a flattened cloth in half along the edge with two end-effectors. The coordinates of four corners $(x,y,z)$ are applied as the reduced states. The performance is evaluated by measuring the alignment of two sets of folded corner positions with a penalty based on the displacement of the cloth from its original position. 
\end{itemize} 

\begin{itemize}
\item
\textit{Cloth Fold Diagonal Pinned Image}:
The objective of the task is to fold the cloth along a specified diagonal of a square cloth. The coordinates of four corners $(x,y,z)$ are applied as the reduced states. A heavy block pins one corner of the cloth on the table. The performance is evaluated by measuring the alignment of bottom-left and top-left corner positions.
\end{itemize}

\begin{itemize}
\item
\textit{Cloth Fold Diagonal Unpinned Image}:
The objective of the task is to fold the cloth along a specified diagonal of a square cloth. The coordinates of four corners $(x,y,z)$ are applied as the reduced states. The cloth is free to move on the tabletop. The performance is evaluated by measuring the alignment of bottom-left and top-left corner positions.
\end{itemize}

We report all the results by a normalized version of the performance score provided by SoftGym to compare with other baselines. The normalized performance at timestep $t$, $s_{t}^{'}$ is calculated by $s_{t}^{'}=(s_{t}-s_{0})/(s_{opt}-s_{0})$, where $s_{opt}$ is the optimal performance at the task. Following the protocol of ~\cite{lin2021softgym, salhotra2022learning}, we use the normalized performance at the end of the episode.

For the training of CNN, we apply a four-layer convolution neural network ({15M}) to project the 32x32x3 RGB image into the length of the reduced state. Unlike other RL algorithms, the performance of DT benefits from scaling the model size~\cite{zheng2022online,lee2022multi}. Therefore, we adopt a DT model with the embedded dimension of 256, 10 hidden layers and 16 heads, which yields around 40M trainable parameters. After generating an action vector, a Tanh function is employed to activate end-effector actions. We optimize our model parameter with the LAMB optimizer~\cite{you2019large}. The learning rate is set to 1e-4 with a weight decay of 1e-4.

We use the 8000 episodes with a horizon of 100 steps collected from the heuristic algorithm for the offline training of privileged policy and CNN encoder. A total of 200k state-observation-action steps are applied to train privileged policy and CNN for each task. For the student policy, it only takes 10k steps for the image-based policy to converge. We implement our network by Pytorch. All the training is conducted on a machine with Intel Core i9-12900KF and Nvidia GeForce GTX 3080Ti GPU.

\begin{table*}[t!]
 \renewcommand\arraystretch{1}
    \centering
     \vspace{4mm}
    \caption{Performance of Robotic Cloth Manipulation Tasks}
    \vspace{-2mm}
   \setlength{\tabcolsep}{2.3mm}{ 
   \begin{tabular}{c c c c c c c c c}
        \hline
        \textbf{tasks} & &\textbf{Expert-state} & \textbf{Teacher-state} & \textbf{TraKDis(ours)}  &\textbf{DMFD} & \textbf{Behavior Cloning} & \textbf{CURL} & \textbf{DT-image}  \\ [0.1ex]

        \hline
     
       \multirow{5}{*}{\textbf{\makecell{Cloth Fold Image}}}       
                            &\multirow{2}{*}{\bm{$\mu \pm \sigma$}} 
                                                 &0.706&0.740&$\textbf{0.583}$ &0.364 & 0.159 &0.139 &0.305 \\
                                                &&$\pm0.159$&$\pm0.111$  &\textbf{$\pm$0.190}&$\pm$0.432 & $\pm$0.080& $\pm$0.312&$\pm$0.207 \\
                            &$\mathbf{25^{th}\%}$&0.637&0.663  &$\textbf{0.467}$ &0.000 &0.111 &0.000 &0.162\\
                            &$\mathbf{median}$   &0.726&0.750  &$\textbf{0.610}$ &0.517 & 0.178 &0.062 & 0.299 \\
                            &$\mathbf{75^{th}\%}$&0.815&0.820  &$\textbf{0.705}$&0.705&0.221  &0.396&0.446\\
        \hline
        \multirow{5}{*}{\textbf{\makecell{Cloth Fold Diagonal \\Pinned Image}}}       
                            &\multirow{2}{*}{\bm{$\mu \pm \sigma$}} 
                                                &0.907& 0.902 &$\textbf{0.916}$ & {0.778} & {0.613} & 0.670 &0.469\\
                            &                   &$\pm$0.009  &$\pm$0.004     &\textbf{$\pm$0.002} &  {$\pm$0.014} & {$\pm$0.008} & $\pm$0.033 &$\pm$0.012  \\
                            &$\mathbf{25^{th}\%}$&0.898& 0.900&$\textbf{0.916}$ &  {0.770}& {0.608} & 0.670 &0.460 \\
                            &$\mathbf{median}$   &0.906&0.901 &$\textbf{0.916}$ &  {0.778} & {0.610} & 0.677  &0.464\\
                            &$\mathbf{75^{th}\%}$&0.915& 0.901&$\textbf{0.917}$ &  {0.789} & {0.615} &  0.683 &0.477\\

            \hline
            \multirow{5}{*}{\textbf{\makecell{Cloth Fold Diagonal \\Unpinned Image}}}     
                            &\multirow{2}{*}{\bm{$\mu \pm \sigma$}} 
                                                &0.926&  0.921&$\textbf{0.944}$ &  {0.861} & {0.663} & 0.827 &0.512 \\
                                               &&$\pm$0.012&$\pm$0.004 &\textbf{$\pm$0.004} & {$\pm$0.004} & {$\pm$0.067}& $\pm$0.034 &$\pm$0.014  \\
                            &$\mathbf{25^{th}\%}$&0.917&0.918 &$\textbf{0.941}$ & {0.859} & { 0.618} & 0.810  &0.505\\
                            &$\mathbf{median}$&0.923& 0.920&$\textbf{0.944}$ & {0.862} & {0.671} &  0.831 &0.510\\
                            &$\mathbf{75^{th}\%}$&0.937&0.923 &$\textbf{0.946}$ &  {0.863}& {0.719} & 0.851 &0.521\\
            \hline
    \end{tabular}}
       \par
         \begin{flushleft}
Comparison Results of Different Model Performances. Expert refers to the oracle with access to the full cloth state and dynamics. The value is obtained from normalized performance defined in the SoftGym environment. The model is obtained from the end of training for each method. We test the model from 100 simulation trials and obtain the mean, standard deviation, 25\% and 75\% percentiles of performance. For the evaluation, we use a fixed value of -1 and 30 for the initial reward-to-go and context length, respectively. We use the same hyperparameters and pre-trained weights from the official DMFD implementation.  {DMFD, Behaviour Cloning, and CURL\mbox{\cite{laskin2020curl}} have a model size of 15M, 36M, and 16M, respectively.}

\end{flushleft}
     \vspace{-18pt}
    \label{table:hanging}
\end{table*}

\begin{figure}[t!]
    \centering
    \includegraphics[width=0.98\columnwidth]{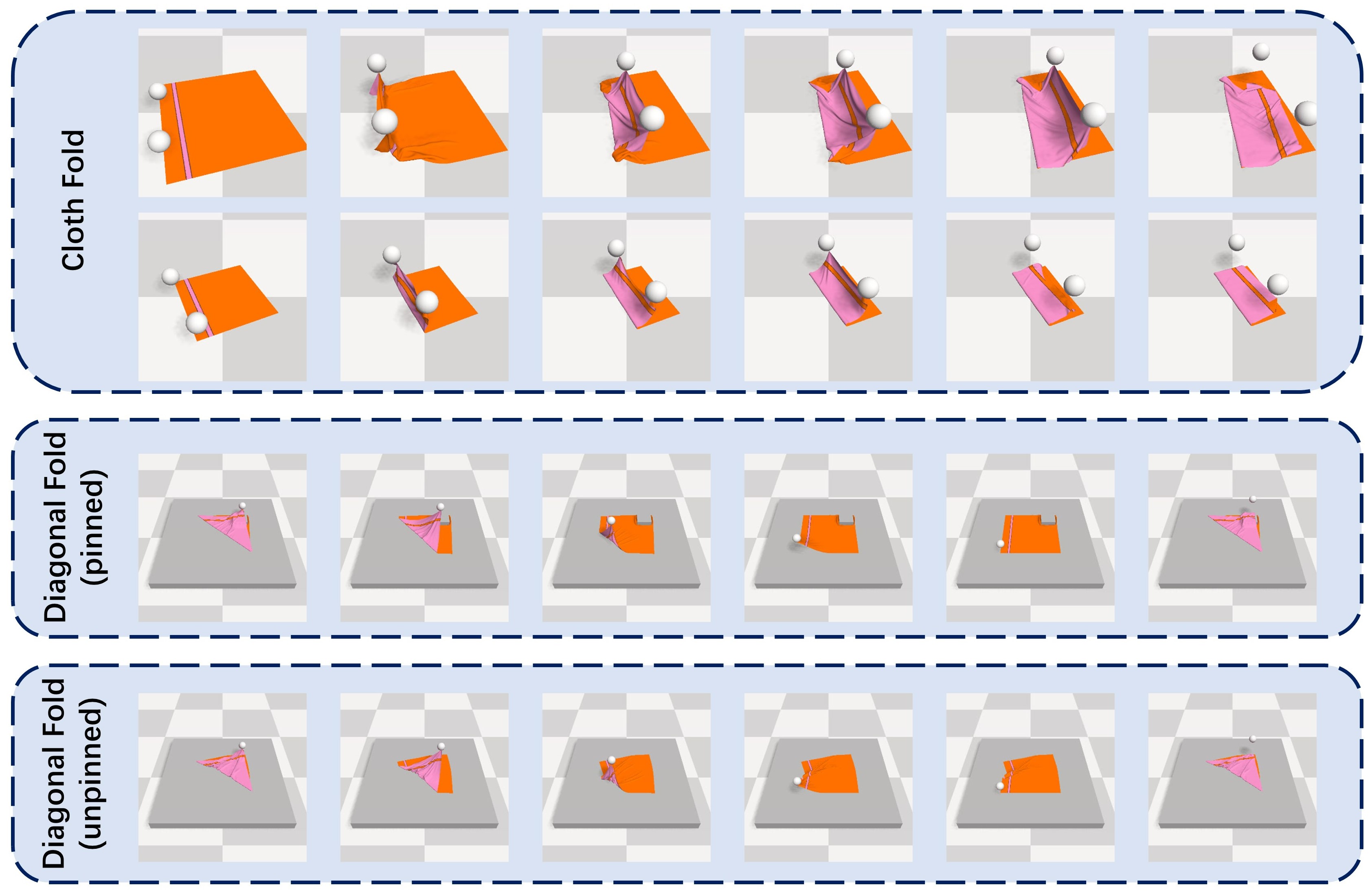}
    \caption{Snapshots of the proposed method. We evaluate our proposed methods by adopting three cloth manipulation tasks. Only image observation is used for the agent to generate trajectory. More demonstration can be found in the multi-media resource}
    \label{fig:sample run}
    \vspace{-15pt}
\end{figure}

\begin{figure}[t!]
    \centering
    \includegraphics[width=0.6\columnwidth]{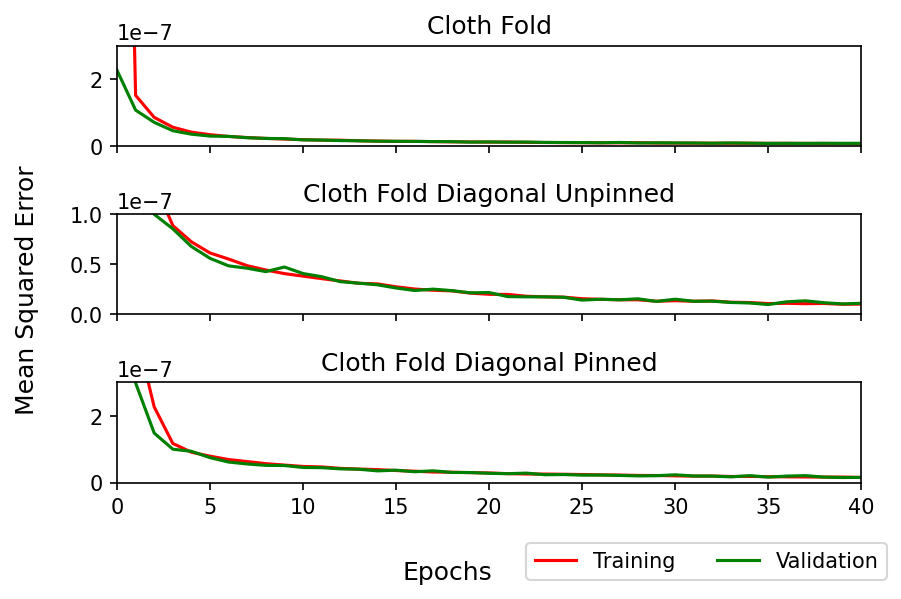}
    \caption{The training and validation errors for CNN encoder in three different tasks}
    \label{fig:CNN evaluation}
    \vspace{-15pt}
\end{figure}

\begin{figure*}[t!]
\centering
\subfigure[ {Cloth Fold}]{
\begin{minipage}[t]{0.325\linewidth}
\centering
\includegraphics[width=2.3in]{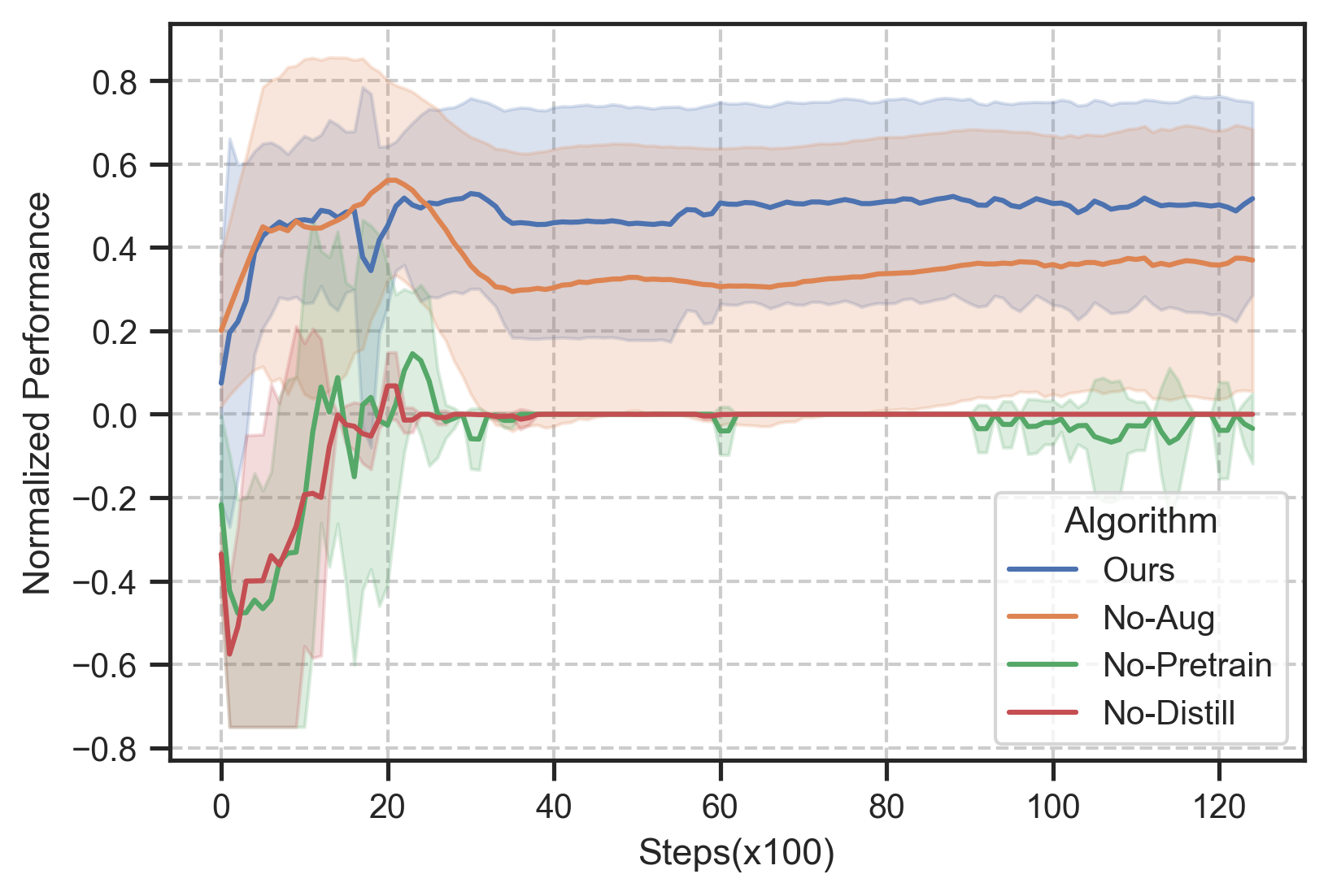}
\end{minipage}%
}%
\subfigure[ {Cloth Diagonal Fold Pinned}]{
\begin{minipage}[t]{0.325\linewidth}
\centering
\includegraphics[width=2.3in]{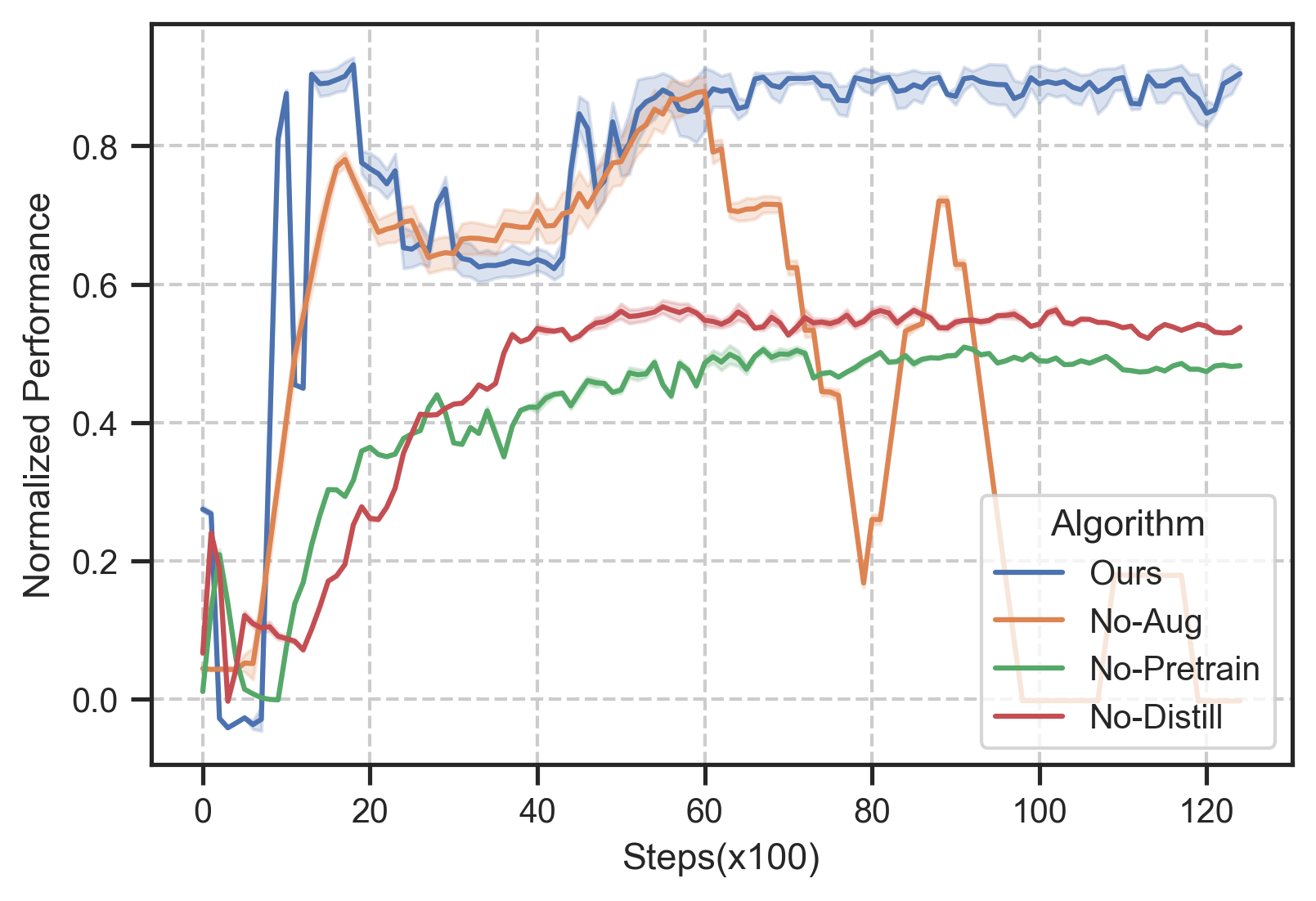}
\end{minipage}%
}%
\subfigure[ {Cloth Diagonal Fold Unpinned}]{
\begin{minipage}[t]{0.325\linewidth}
\centering
\includegraphics[width=2.3in]{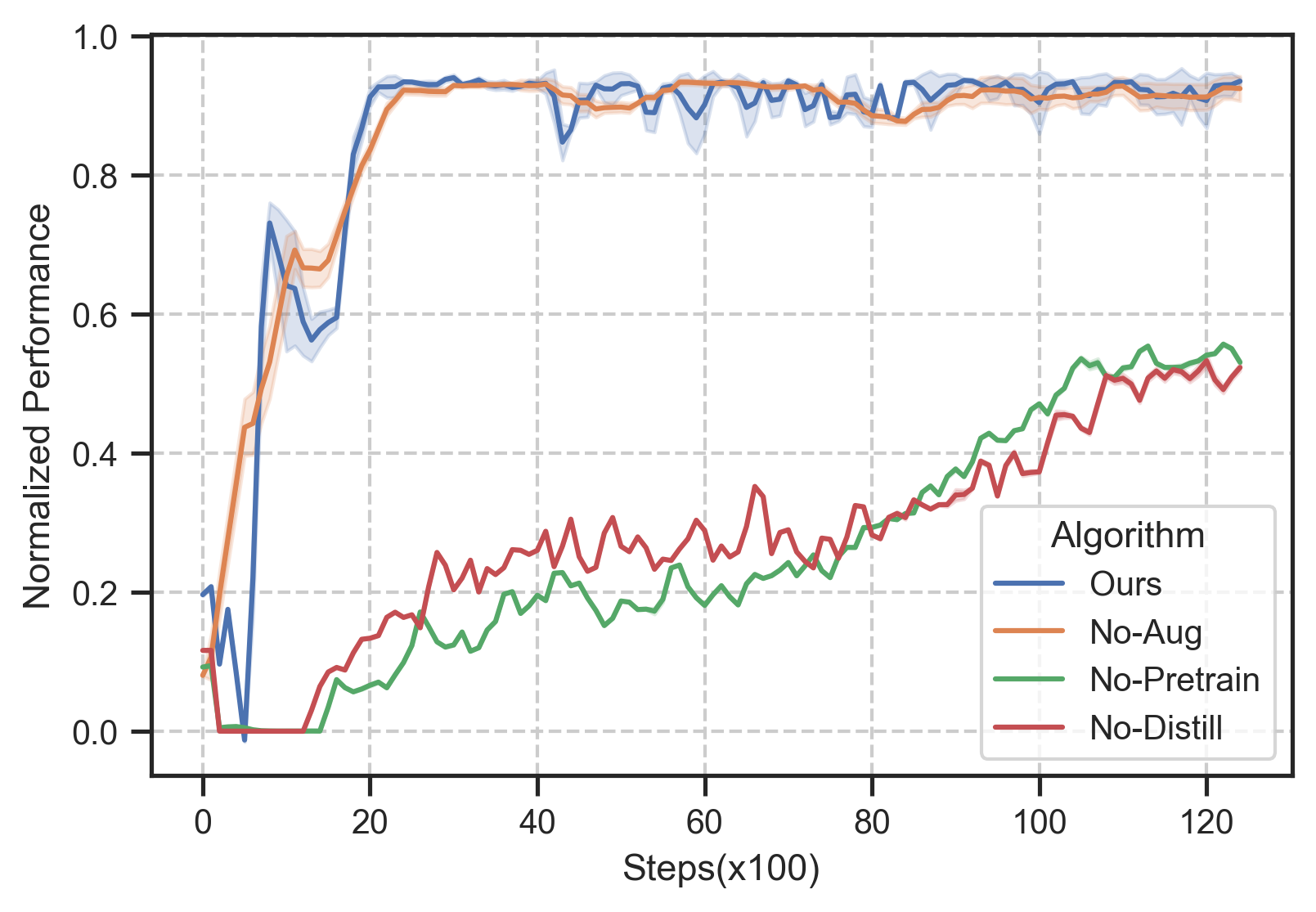}
\end{minipage}
}%
\vspace{-8pt}
\centering
\caption{Learning Curves of Ablation Studies. We attain the normalized performance from five random running seeds. We draw the mean performance as the solid lines. Shaded areas stand for the standard deviation of normalized performance. Thanks to the weight initialization, the student model can converge within 60 epochs for all three tasks. }
 \label{fig:abalation_study}
 \vspace{-10pt}
\end{figure*}

\subsection{Comparison Experiment}
We visualize our pre-trained CNN encoder's training and validation loss in Fig. \ref{fig:CNN evaluation}. All the training procedures for three tasks require 20 epochs to be converged.
We evaluate our proposed method against several state-of-the-art methods for cloth 
manipulation to show the advantages of our method in visual robotic cloth manipulation tasks.

\begin{itemize}
\item
\textbf{Teacher}: The teacher DT model trained with privileged cloth states information.

\item
\textbf{DMFD}: image-based state-of-the-art method that combines expert demonstration and RL for deformable object manipulation from~\cite{salhotra2022learning}.

\item
\textbf{Behaviour Cloning}: a behaviour cloning policy that takes image and action pairs for policy training to imitate or replicate behaviours demonstrated by the expert.

\item
\textbf{CURL}: a state-of-the-art image-based RL algorithm that utilizes contrastive representation for RL agent learning.

\item
    {\textbf{DT-image}: We directly use the RGB image as the input to train the agent. The CNN encoder for state estimation is not applied for training}

\end{itemize}

We use the official implementation from SoftGym and DMFD. All the hyperparameters are kept the same without any modification Table I shows all the results of the comparison experiments. As illustrated in Table I, our methods outperform other baselines in all three image-based tasks. A detailed discussion will be presented in section IV-F.

\subsection{Ablation Studies}

We conduct ablation studies to verify the necessity of each component of our proposed method. 
\begin{itemize}
\item
\textbf{Ours}:
We pre-train a CNN encoder \textbf{with} the image augmentation technique, copy the weight from the teacher agent, and imitate the teacher policy.

\item
\textbf{No-Aug}:
We pre-train a CNN encoder \textbf{without} image augmentation technique, copy the weight from the teacher agent, and imitate the teacher policy.

\item
\textbf{No-Pretrain}:
We pre-train a CNN encoder with image augmentation technique, \textbf{without} copy weight, and imitate the teacher policy.

\item
     {\textbf{No-Distill}: We remove the distillation part by using the decision transformer and the CNN encoder to imitate the heuristic expert demonstrations.
    }
    
\end{itemize}

We draw the learning curves of several variants of our proposed methods as shown in Fig. \ref{fig:abalation_study}. The results show the effectiveness of every component of our proposed methods. A detailed discussion will be presented in Section IV-F.

\subsection{Robustness}
To test the robustness of our proposed method, we evaluate our model in a more challenging environment. In this experiment, We introduce a Gaussian noise $\epsilon$ with a mean of zero and standard deviation $\sigma=1e-3$ to the estimated states $S^{'}$ for the model training.  This aims to test the model's robustness when the CNN has an inaccurate estimation. We use the relatively difficult Cloth Fold task as the case study. The student model is trained for 60 episodes according to the training curves from Fig.\ref{fig:abalation_study} (a).

\begin{figure}[t!]
\centering
\vspace{4mm}
\subfigure[]{
\begin{minipage}[t]{0.525\linewidth}
\centering
\includegraphics[width=1\columnwidth]{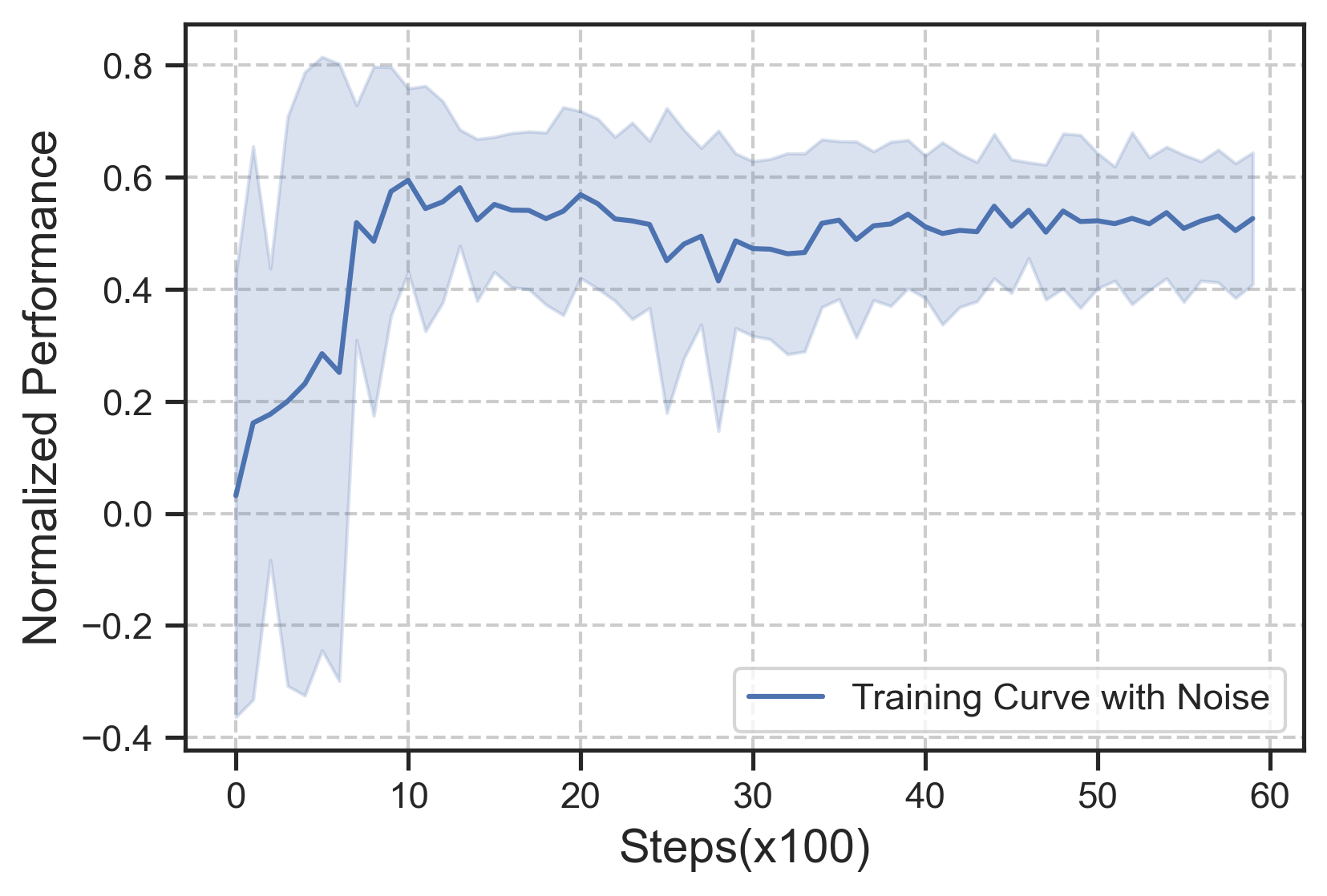}
\end{minipage}}
\centering
\subfigure[]{
\begin{minipage}[t]{0.525\linewidth}
\centering
\includegraphics[width=1\columnwidth]{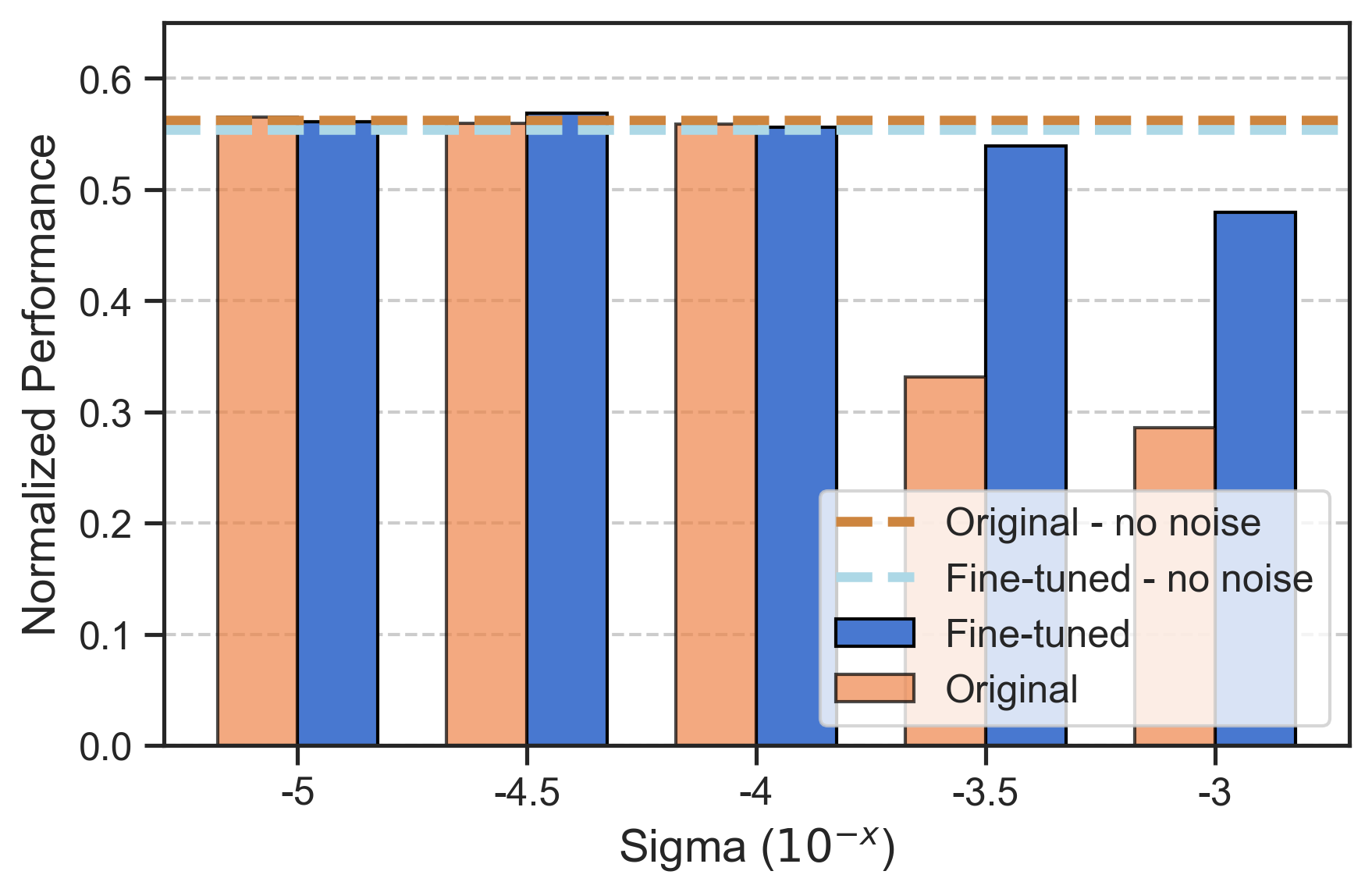}
\end{minipage}%
}

\centering
\caption{(a) Training curves with noise added. We train our student model with added Gaussian noise. It can be observed that our model can adapt to the added noise and increase its robustness.  
(b) Comparison of models. The model trained with added noise has a higher tolerance to added noise. Dashed lines represent the performance where no noise is added. We obtain the result from a mean value of 10 simulation trials.}
 \label{fig:robustness}
 \vspace{-10pt}
\end{figure}

The results are demonstrated in Fig. \ref{fig:robustness}. It can be observed that added noise does not degrade the model training and final performance. We further conduct a comparison experiment of models trained with added noise and without noise.
This is achieved by evaluating the model by adding noise that ranges from $\sigma=1e-5$ to $\sigma=1e-3$ in Fig. \ref{fig:robustness} (b). The results indicate that the model fine-tuned with noise exhibits a higher tolerance for inaccurate state estimation with more significant noise added.

$$
S^{'}_{noise} = S^{'} + \epsilon, \quad \epsilon\sim \mathcal{N}(\mu,\,\sigma^{2}) 
$$

\subsection{Real-world Experiment}
\begin{figure}[t!]
    \centering
    \vspace{4mm}
    \includegraphics[width=0.98\columnwidth]{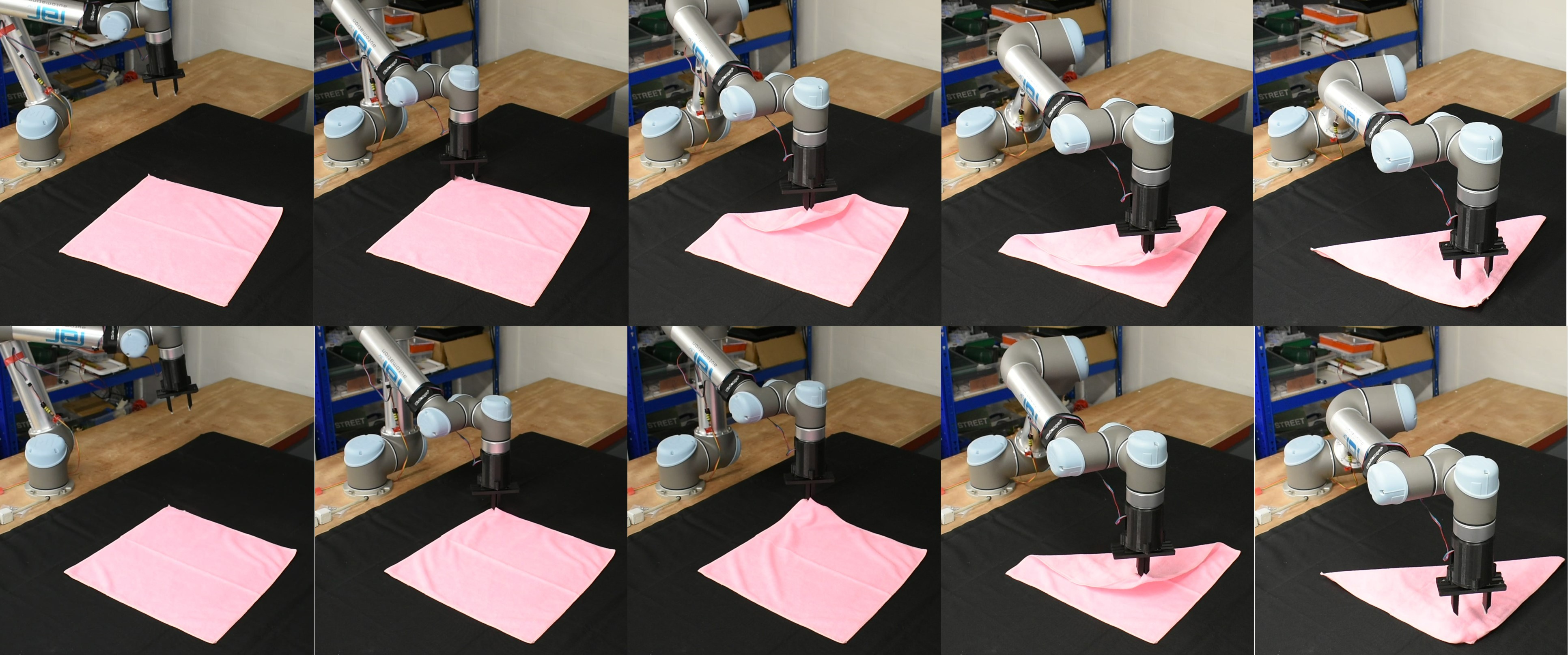}
    \caption{Real-world demonstration for cloth diagonal unpinned task}
    \label{fig:real demo}
    \vspace{-10pt}
\end{figure}

In the real world, we perform the Cloth Fold Unpinned task to test the effectiveness of our proposed method. This is implemented by applying a UR5 robot and a customized parallel gripper, as shown in Fig.\ref{fig:real demo}. We rotate the target cloth to different angles to test the performance.
One Intel RealSense D435i RGB-D camera is fixed above the scene during the experiment to capture the images. We follow a similar processing procedure in~\cite{salhotra2022learning} to pre-process our image input to ensure the robustness of our system. No fine-tuning or re-training is performed to fit the real-world setting. Our method is evaluated on a total of ten rollouts.
For the simulation, we obtain a mean result of 0.940. For the real world, we obtain a performance of 0.895.
More demonstrations can be found in the multi-media resource.

\subsection{Discussion}

According to the results presented in Table I, our state-based agent demonstrates comparable or superior performance compared to the expert policy. In the case of the image-based environment, \textit{TraKDis} consistently achieves the highest performance across all three tasks. More specifically, our model achieves a performance of 0.583, which corresponds to 78\% of the teacher's performance in the challenging cloth fold task. In addition, our approach achieves performances of 0.916 and 0.944 in the cloth diagonal pinned and unpinned fold tasks, respectively, surpassing both the expert and teacher's performances.  
Compared to other baselines, \textit{TraKDis} exhibits an improved performance across three tasks, with enhancements of 21.9\%,  {13.8\%}, and  {8.3\%} observed in these respective tasks. Moreover, our algorithm exhibits the lowest standard deviation in terms of model robustness, indicating its reliability and consistency.

By utilizing the image augmentation technique, our pre-trained CNN can employ a more robust state estimation, which stabilizes the training of student policy, as shown in the ablation study.
In addition, by initializing the weight from the privileged agent, we significantly improve the training efficiency and performance. The training of our student policy only takes 60 episodes (6000 steps) for the policy to be converged, which is only 0.6\% of the total training steps reported in~\cite{salhotra2022learning}. 
\hl{However, due to the absence of temporal information in the CNN encoder as a state estimator, errors in estimation and inaccuracy regarding action prediction may occur. While this could result in an error accumulation problem for MDP-based methods, we mitigate this issue by using the DT model, adapting the estimated state in a temporal sequence level.}
Besides, compared to the ablation variants, the weight copy also provides a good initialization for the student policy training. Without the leverage of the teacher's weight, the model exhibits either a slow performance improvement or performance degradation as shown in the \textbf{No-Pretrain} and  {\textbf{No-Distill}} ablation variants in Fig.\ref{fig:abalation_study}.

We further evaluate our model in a challenging noisy environment. As shown in Fig.\ref{fig:robustness}, It should be observed that our model can adapt to the added noise and converge within 60 epochs.
Although adding noise fails to improve the model's performance, the added noise can be beneficial to the model's robustness as the standard deviation decreases with more training steps, shown in Fig.\ref{fig:robustness}.
Results also indicate that by re-training the student model with noise added, our model can obtain a higher tolerance to inaccurate state estimation. Real-world experiments are also conducted to show the effectiveness of our proposed method in the real-world setting.

\section{Conclusions}
In this work, we present a novel approach named \textit{TraKDis} for the learning of visual RL policy for cloth manipulation tasks. 
While direct training from images is extremely difficult, by leveraging knowledge from the model with access to privileged information, \textit{TraKDis} tries to solve this problem with a one-to-one novel KD approach that leverages weight copy and input state estimation.  {Although the transformer model has more parameters than MLP-based methods, its sequence modelling capability, considering historical state information for each action, enhances robustness and efficiency in distillation tasks.}
A comparison experiment is conducted to test the performance respect to state-of-the-art algorithms in three different tasks. The experiment results indicate that \textit{TraKDis} shows better performance in all three tasks. Specifically, our proposed approach demonstrates an increased performance of 21.9\%,  {13.8\%}, and  {8.3\%} in these three tasks.
Ablation experiments and robustness experiments are also conducted for a more in-depth analysis of our proposed method.  
To validate our approach further, a real-world experiment is also conducted.

Despite the benefits of the approach, there are some limitations. For instance, our work collects 8000 episodes (80k state-action-reward samples) for privileged policy training. Future work can focus on improving the data efficiency since the current agent requires a large state and action transition dataset for offline training. Besides, we observe that there is still a 22\% performance gap for the distillation of the cloth folding task. \hl{Moreover, in contrast to conventional RL approaches, \textit{TraKDis} involves a larger model size. This could potentially lead to issues related to operational speed and memory utilization. Future research could focus on optimizing and reducing the overall size of the model.}
 {With recent progress in visual deep RL \mbox{\cite{shah2021rrl,wang2022vrl3}}, future work can be devoted towards integrating pre-trained encoder with large-scale image datasets and our distillation approach for more challenging tasks such as complex-shaped cloth (e.g. shirts and trousers) manipulation and multi-goal cloth manipulation tasks.
}

\addtolength{\textheight}{-1cm}   


\bibliographystyle{IEEEtran}
\bibliography{main}   

\begin{thebibliography}{10}
\providecommand{\url}[1]{#1}
\csname url@rmstyle\endcsname
\providecommand{\newblock}{\relax}
\providecommand{\bibinfo}[2]{#2}
\providecommand\BIBentrySTDinterwordspacing{\spaceskip=0pt\relax}
\providecommand\BIBentryALTinterwordstretchfactor{4}
\providecommand\BIBentryALTinterwordspacing{\spaceskip=\fontdimen2\font plus
\BIBentryALTinterwordstretchfactor\fontdimen3\font minus
  \fontdimen4\font\relax}
\providecommand\BIBforeignlanguage[2]{{%
\expandafter\ifx\csname l@#1\endcsname\relax
\typeout{** WARNING: IEEEtran.bst: No hyphenation pattern has been}%
\typeout{** loaded for the language `#1'. Using the pattern for}%
\typeout{** the default language instead.}%
\else
\language=\csname l@#1\endcsname
\fi
#2}}

\bibitem{doi:10.1126/scirobotics.abo7229}
\BIBentryALTinterwordspacing
J.~Borràs, ``Effective grasping enables successful robot-assisted dressing,''
  \emph{Science Robotics}, vol.~7, no.~65, p. eabo7229, 2022. [Online].
  Available: \url{https://www.science.org/doi/abs/10.1126/scirobotics.abo7229}
\BIBentrySTDinterwordspacing

\bibitem{matas2018sim}
J.~Matas, S.~James, and A.~J. Davison, ``Sim-to-real reinforcement learning for
  deformable object manipulation,'' in \emph{Conference on Robot
  Learning}.\hskip 1em plus 0.5em minus 0.4em\relax PMLR, 2018, pp. 734--743.

\bibitem{lin2021softgym}
X.~Lin, Y.~Wang, J.~Olkin, and D.~Held, ``Softgym: Benchmarking deep
  reinforcement learning for deformable object manipulation,'' in
  \emph{Conference on Robot Learning}.\hskip 1em plus 0.5em minus 0.4em\relax
  PMLR, 2021, pp. 432--448.

\bibitem{9196659}
R.~Jangir, G.~Alenyà, and C.~Torras, ``Dynamic cloth manipulation with deep
  reinforcement learning,'' in \emph{2020 IEEE International Conference on
  Robotics and Automation (ICRA)}, 2020, pp. 4630--4636.

\bibitem{wu2019learning}
Y.~Wu, W.~Yan, T.~Kurutach, L.~Pinto, and P.~Abbeel, ``Learning to manipulate
  deformable objects without demonstrations,'' \emph{arXiv preprint
  arXiv:1910.13439}, 2019.

\bibitem{salhotra2022learning}
G.~Salhotra, I.-C.~A. Liu, M.~Dominguez-Kuhne, and G.~S. Sukhatme, ``Learning
  deformable object manipulation from expert demonstrations,'' \emph{IEEE
  Robotics and Automation Letters}, vol.~7, no.~4, pp. 8775--8782, 2022.

\bibitem{levine2016end}
S.~Levine, C.~Finn, T.~Darrell, and P.~Abbeel, ``End-to-end training of deep
  visuomotor policies,'' \emph{The Journal of Machine Learning Research},
  vol.~17, no.~1, pp. 1334--1373, 2016.

\bibitem{levine2018learning}
S.~Levine, P.~Pastor, A.~Krizhevsky, J.~Ibarz, and D.~Quillen, ``Learning
  hand-eye coordination for robotic grasping with deep learning and large-scale
  data collection,'' \emph{The International journal of robotics research},
  vol.~37, no. 4-5, pp. 421--436, 2018.

\bibitem{pinto2017asymmetric}
L.~Pinto, M.~Andrychowicz, P.~Welinder, W.~Zaremba, and P.~Abbeel, ``Asymmetric
  actor critic for image-based robot learning,'' \emph{arXiv preprint
  arXiv:1710.06542}, 2017.

\bibitem{liu2022distilling}
I.-C.~A. Liu, S.~Uppal, G.~S. Sukhatme, J.~J. Lim, P.~Englert, and Y.~Lee,
  ``Distilling motion planner augmented policies into visual control policies
  for robot manipulation,'' in \emph{Conference on Robot Learning}.\hskip 1em
  plus 0.5em minus 0.4em\relax PMLR, 2022, pp. 641--650.

\bibitem{10160946}
W.~Ding, N.~Majcherczyk, M.~Deshpande, X.~Qi, D.~Zhao, R.~Madhivanan, and
  A.~Sen, ``Learning to view: Decision transformers for active object
  detection,'' in \emph{2023 IEEE International Conference on Robotics and
  Automation (ICRA)}, 2023, pp. 7140--7146.

\bibitem{ramakrishnan2021exploration}
S.~K. Ramakrishnan, D.~Jayaraman, and K.~Grauman, ``An exploration of embodied
  visual exploration,'' \emph{International Journal of Computer Vision}, vol.
  129, pp. 1616--1649, 2021.

\bibitem{ding2023learning}
W.~Ding, N.~Majcherczyk, M.~Deshpande, X.~Qi, D.~Zhao, R.~Madhivanan, and
  A.~Sen, ``Learning to view: Decision transformers for active object
  detection,'' \emph{arXiv preprint arXiv:2301.09544}, 2023.

\bibitem{humeau2019poly}
S.~Humeau, K.~Shuster, M.-A. Lachaux, and J.~Weston, ``Poly-encoders:
  Transformer architectures and pre-training strategies for fast and accurate
  multi-sentence scoring,'' \emph{arXiv preprint arXiv:1905.01969}, 2019.

\bibitem{zhu2023transfer}
Z.~Zhu, K.~Lin, A.~K. Jain, and J.~Zhou, ``Transfer learning in deep
  reinforcement learning: A survey,'' \emph{IEEE Transactions on Pattern
  Analysis and Machine Intelligence}, 2023.

\bibitem{9097275}
J.~Borràs, G.~Alenyà, and C.~Torras, ``A grasping-centered analysis for cloth
  manipulation,'' \emph{IEEE Transactions on Robotics}, vol.~36, no.~3, pp.
  924--936, 2020.

\bibitem{8957044}
I.~Garcia-Camacho, M.~Lippi, M.~C. Welle, H.~Yin, R.~Antonova, A.~Varava,
  J.~Borras, C.~Torras, A.~Marino, G.~Alenyà, and D.~Kragic, ``Benchmarking
  bimanual cloth manipulation,'' \emph{IEEE Robotics and Automation Letters},
  vol.~5, no.~2, pp. 1111--1118, 2020.

\bibitem{seita_bedmake_2019}
D.~Seita, N.~Jamali, M.~Laskey, A.~K. Tanwani, R.~Berenstein, P.~Baskaran,
  S.~Iba, J.~Canny, and K.~Goldberg, ``{Deep Transfer Learning of Pick Points
  on Fabric for Robot Bed-Making},'' in \emph{International Symposium on
  Robotics Research (ISRR)}, 2019.

\bibitem{qian2020cloth}
J.~Qian, T.~Weng, L.~Zhang, B.~Okorn, and D.~Held, ``Cloth region segmentation
  for robust grasp selection,'' in \emph{2020 IEEE/RSJ International Conference
  on Intelligent Robots and Systems (IROS)}.\hskip 1em plus 0.5em minus
  0.4em\relax IEEE, 2020, pp. 9553--9560.

\bibitem{ha2021flingbot}
H.~Ha and S.~Song, ``Flingbot: The unreasonable effectiveness of dynamic
  manipulation for cloth unfolding,'' in \emph{Conference on Robotic Learning
  (CoRL)}, 2021.

\bibitem{xu2022dextairity}
Z.~Xu, C.~Chi, B.~Burchfiel, E.~Cousineau, S.~Feng, and S.~Song, ``Dextairity:
  Deformable manipulation can be a breeze,'' in \emph{Proceedings of Robotics:
  Science and Systems (RSS)}, 2022.

\bibitem{chen2023learning}
W.~Chen, D.~Lee, D.~Chappell, and N.~Rojas, ``Learning to grasp clothing
  structural regions for garment manipulation tasks,'' \emph{arXiv preprint
  arXiv:2306.14553}, 2023.

\bibitem{Wang2023One}
Y.~Wang, Z.~Sun, Z.~Erickson, and D.~Held, ``One policy to dress them all:
  Learning to dress people with diverse poses and garments,'' in
  \emph{Robotics: Science and Systems (RSS)}, 2023.

\bibitem{laskin2020curl}
M.~Laskin, A.~Srinivas, and P.~Abbeel, ``Curl: Contrastive unsupervised
  representations for reinforcement learning,'' in \emph{International
  Conference on Machine Learning}.\hskip 1em plus 0.5em minus 0.4em\relax PMLR,
  2020, pp. 5639--5650.

\bibitem{hafner2019learning}
D.~Hafner, T.~Lillicrap, I.~Fischer, R.~Villegas, D.~Ha, H.~Lee, and
  J.~Davidson, ``Learning latent dynamics for planning from pixels,'' in
  \emph{International conference on machine learning}.\hskip 1em plus 0.5em
  minus 0.4em\relax PMLR, 2019, pp. 2555--2565.

\bibitem{vaswani2017attention}
A.~Vaswani, N.~Shazeer, N.~Parmar, J.~Uszkoreit, L.~Jones, A.~N. Gomez,
  {\L}.~Kaiser, and I.~Polosukhin, ``Attention is all you need,''
  \emph{Advances in neural information processing systems}, vol.~30, 2017.

\bibitem{dosovitskiy2020image}
A.~Dosovitskiy, L.~Beyer, A.~Kolesnikov, D.~Weissenborn, X.~Zhai,
  T.~Unterthiner, M.~Dehghani, M.~Minderer, G.~Heigold, S.~Gelly,
  \emph{et~al.}, ``An image is worth 16x16 words: Transformers for image
  recognition at scale,'' \emph{arXiv preprint arXiv:2010.11929}, 2020.

\bibitem{chen2021decision}
L.~Chen, K.~Lu, A.~Rajeswaran, K.~Lee, A.~Grover, M.~Laskin, P.~Abbeel,
  A.~Srinivas, and I.~Mordatch, ``Decision transformer: Reinforcement learning
  via sequence modeling,'' \emph{Advances in neural information processing
  systems}, vol.~34, pp. 15\,084--15\,097, 2021.

\bibitem{zheng2022online}
Q.~Zheng, A.~Zhang, and A.~Grover, ``Online decision transformer,'' in
  \emph{International Conference on Machine Learning}.\hskip 1em plus 0.5em
  minus 0.4em\relax PMLR, 2022, pp. 27\,042--27\,059.

\bibitem{lee2022multi}
K.-H. Lee, O.~Nachum, M.~S. Yang, L.~Lee, D.~Freeman, S.~Guadarrama,
  I.~Fischer, W.~Xu, E.~Jang, H.~Michalewski, \emph{et~al.}, ``Multi-game
  decision transformers,'' \emph{Advances in Neural Information Processing
  Systems}, vol.~35, pp. 27\,921--27\,936, 2022.

\bibitem{hinton2015distilling}
G.~Hinton, O.~Vinyals, and J.~Dean, ``Distilling the knowledge in a neural
  network,'' \emph{stat}, vol. 1050, p.~9, 2015.

\bibitem{kumar2021rma}
A.~Kumar, Z.~Fu, D.~Pathak, and J.~Malik, ``Rma: Rapid motor adaptation for
  legged robots,'' 2021.

\bibitem{lee2020learning}
J.~Lee, J.~Hwangbo, L.~Wellhausen, V.~Koltun, and M.~Hutter, ``Learning
  quadrupedal locomotion over challenging terrain,'' \emph{Science robotics},
  vol.~5, no.~47, p. eabc5986, 2020.

\bibitem{janner2021offline}
M.~Janner, Q.~Li, and S.~Levine, ``Offline reinforcement learning as one big
  sequence modeling problem,'' \emph{Advances in neural information processing
  systems}, vol.~34, pp. 1273--1286, 2021.

\bibitem{laskin2020reinforcement}
M.~Laskin, K.~Lee, A.~Stooke, L.~Pinto, P.~Abbeel, and A.~Srinivas,
  ``Reinforcement learning with augmented data,'' \emph{Advances in neural
  information processing systems}, vol.~33, pp. 19\,884--19\,895, 2020.

\bibitem{devlin2018bert}
J.~Devlin, M.-W. Chang, K.~Lee, and K.~Toutanova, ``Bert: Pre-training of deep
  bidirectional transformers for language understanding,'' \emph{arXiv preprint
  arXiv:1810.04805}, 2018.

\bibitem{you2019large}
Y.~You, J.~Li, S.~Reddi, J.~Hseu, S.~Kumar, S.~Bhojanapalli, X.~Song,
  J.~Demmel, K.~Keutzer, and C.-J. Hsieh, ``Large batch optimization for deep
  learning: Training bert in 76 minutes,'' \emph{arXiv preprint
  arXiv:1904.00962}, 2019.

\bibitem{shah2021rrl}
R.~M. Shah and V.~Kumar, ``Rrl: Resnet as representation for reinforcement
  learning,'' in \emph{International Conference on Machine Learning}.\hskip 1em
  plus 0.5em minus 0.4em\relax PMLR, 2021, pp. 9465--9476.

\bibitem{wang2022vrl3}
C.~Wang, X.~Luo, K.~Ross, and D.~Li, ``Vrl3: A data-driven framework for visual
  deep reinforcement learning,'' \emph{Advances in Neural Information
  Processing Systems}, vol.~35, pp. 32\,974--32\,988, 2022.

\end{thebibliography}

\end{document}